\documentclass{article}
\usepackage{spconf,amsmath,epsfig}
\usepackage{subfig}
\usepackage{algorithm}%
\usepackage{algorithmicx}%
\usepackage{algpseudocode}%
\usepackage{array}
\usepackage{wrapfig}

\title{PD-Seg: Population Disaggregation Using Deep Segmentation Networks For Improved Built Settlement Mask}
\name{M. A. Rahman,
 M. A. Waseem,
 Z. Khalid,
 M. Tahir,
 M. Uppal\thanks{
This work was supported under the Grand Challenge Fund of the Higher
Education Commission, Pakistan (Grant Number: GCF-521).}}
\address{Department of Electrical Engineering, Lahore University of Management Sciences, Lahore 54792, Pakistan}
\begin{document}
%
\maketitle
\begin{abstract}
Any policy-level decision-making procedure and academic research involving the optimum use of resources for development and planning initiatives depends on accurate population density statistics. The current cutting-edge datasets offered by WorldPop and Meta do not succeed in achieving this aim for developing nations like Pakistan; the inputs to their algorithms provide flawed estimates that fail to capture the spatial and land-use dynamics. In order to precisely estimate population counts at a resolution of 30 meters by 30 meters, we use an accurate built settlement mask obtained using deep segmentation networks and satellite imagery. The Points of Interest (POI) data is also used to exclude non-residential areas.
\end{abstract}
\section{Introduction}
\label{sec:intro}
For various decision-making processes and development initiatives, such as urban growth, infectious disease containment, evacuation planning, risk management and conservation planning, accurate population density data is essential. Census data disaggregation using survey-based methods lacks the precision needed for these applications and is seldom done because of the time and expense requirements~\cite{robinson2017deep}. 

There have been several attempts in the literature to accurately disaggregate census data. Azar et al use remotely sensed data combined with a likelihood layer to disaggregate census data at 100 meter resolution \cite{azar2013generation}. Linard at al use land-classification and settlement points for disaggregation, also at a resolution of 100 meter \cite{linard2010high}.  However the existing worldwide population grids contain flaws that render them useless \cite{freire2015combining}. Even cutting-edge datasets like WorldPop (100m x 100m) and Meta (30m x 30m) have several flaws, particularly in the case of developing nations like Pakistan where high-quality urban data is not readily accessible for public use \cite{tiecke2017mapping,sorichetta2015high}. For example, both Meta and WorldPop have disaggregated the population based on 2010 census estimates (provided by Demobase), and therefore do not accurately capture the current dynamics of 2023. Additionally, these projections are made at the tehsil level, the second administrative level out of a total of five levels. Furthermore, WorldPop and Meta both utilize low-quality covariates and  methadologies to disaggregate the population, which lead to errors. Meta equally disaggregates the population estimates across built-up tiles for each tehsil while the WorldPop dataset, even the constrained one, displays high population counts in physically desolate regions.

The following contributions address all of the aforementioned issues:
\begin{enumerate}
\item Disaggregation based on the most recent (2017) publicly available census data at the second-highest resolution i.e, fourth administrative level.
\item Accurate built-up mask developed using deep segmentation networks and satellite imagery. Built-up proportions per tile are used to determine population density and Points of Interest (POI) data is used to remove non-residential regions.
\end{enumerate}
The rest of the paper is structured as follows: Methodology (section 2) explains the built-up mask and POI based disaggreagtion technique, Evaluation and Analysis (section 3) discusses and compares our results with state-of-the-art approaches, and Conclusion (section 4) provides the concluding remarks.

\begin{figure*}[t]
\centering
\includegraphics[width = \textwidth]{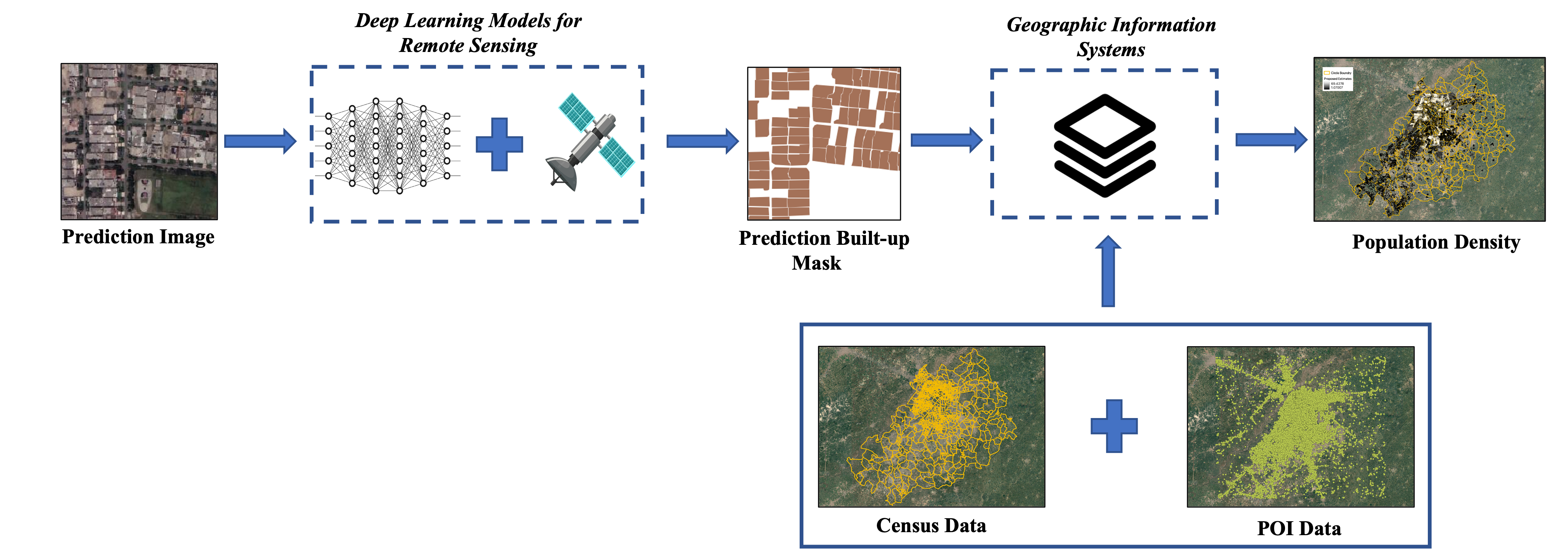}
\caption{Proposed pipeline for processing data. To create a population density map, we first get the built-up estimates using deep learning models using satellite imagery and then combine them with census and POI data.}
\label{Fig: 1}
\end{figure*}


\section{Methodology}
\label{methodology}
As previously mentioned, old low-resolution population aggregates and unreliable constructed settlement masks that support naïve disaggregation methods result in sub-par population density maps for Pakistan. We set out to determine an accurate population density map of Lahore using latest census data, satellite imagery and POI.
\subsection{Census Data}
We obtain the most recent 2017 census data from the PBS website and convert it into GIS vector files up to the circle, or the forth administrative level, as part of the technique shown in Figure \ref{Fig: 1}. The Lahore district is divided into 7 tehsils, 184 charges, 867 circles and 6,764 blocks. The city has a population of around 11.13 million, with approximately 1,744,755 households \cite{bib29, bib30}. A higher administrative level for disaggreagtion significantly improves the density estimates due to a finer resolution of census data.
\subsection{Segmentation Network}
The built-up area prediction masks obtained through the deep semantic segmentation model are used to disaggregate the population counts into 30m × 30m tiles. The deep network is built on the DeepLabV3+ architecture with a dilated ResNet encoder \cite{chen2018encoder}, that is trained using a Dice Loss on manually annotated datasets of various parts of Lahore, Pakistan. We train the model for 80 epochs using an 80-20 Train-Val split and a 8-batch size \cite{waseem2022estimating}. We use Google Earth satellite imagery at a fine resolution of 20 zoom level (about 0.3 meters per pixel) to create high-quality constructed settlement masks. 
\subsection{Disaggregation}
After determining built-up regions, we exclude the tiles containing POI using a mask, denoted by $T_i$ for $i$-th tile as
\setlength{\arraycolsep}{0.0em}
\begin{eqnarray}
T_{i} = \begin{cases}
    0, &  \exists \: j , \: B(R,POI_{ij}) \geq P \: \\
    1,   & \text{otherwise}
\end{cases}
\end{eqnarray}
for $i = 1,2,...N$, $R$ is the radius value, $POI$ is the Points of Interest dataset, $B$ is the buffer and $P$ is the number of points. The algorithm draws a buffer $B$ of radius $R$ around a point in $POI$. If $B$ contains points greater than or equal to $P$, $T_i$ is removed. This method increases the probability of removing only those tiles that cover non-residential built up areas. Individual POI can be located in between residential areas which are less likely to be removed using this method. For this purpose, $R$ is set to 500 meters and $P$ is set to 5. The circle-level population is then divided among the remaining residential tiles by weighting each tile according to the quantity of built-up pixels existing in that tile, as illustrated below.
\setlength{\arraycolsep}{0.0em}
\begin{eqnarray}
P_{\rm tile} = \frac{\sum\limits_{i=1}^{N_t} f(i)}{\sum\limits_{j=1}^{N_c} f(j)}
\times P_{\rm circle}\;\:\:\:\forall\, \rm tile \in \rm circle,
\end{eqnarray}
\noindent where $P_{\rm circle}$ is the population of the circle to which the tile belongs, $N_t$ represents the total number of pixels in the tile, $N_c$ represents the total number of pixels in the circle and $f$ is a function which takes pixel position as input and returns 1 if it belongs to built-up class and 0 otherwise. Using this approach, we are able to concentrate a higher density into the tiles with higher built-up proportions while simultaneously excluding the unbuilt tiles/regions.
\begin{figure*}[!ht]
\centering
\subfloat[]{
  \includegraphics[width=\columnwidth, height = 60mm,keepaspectratio]{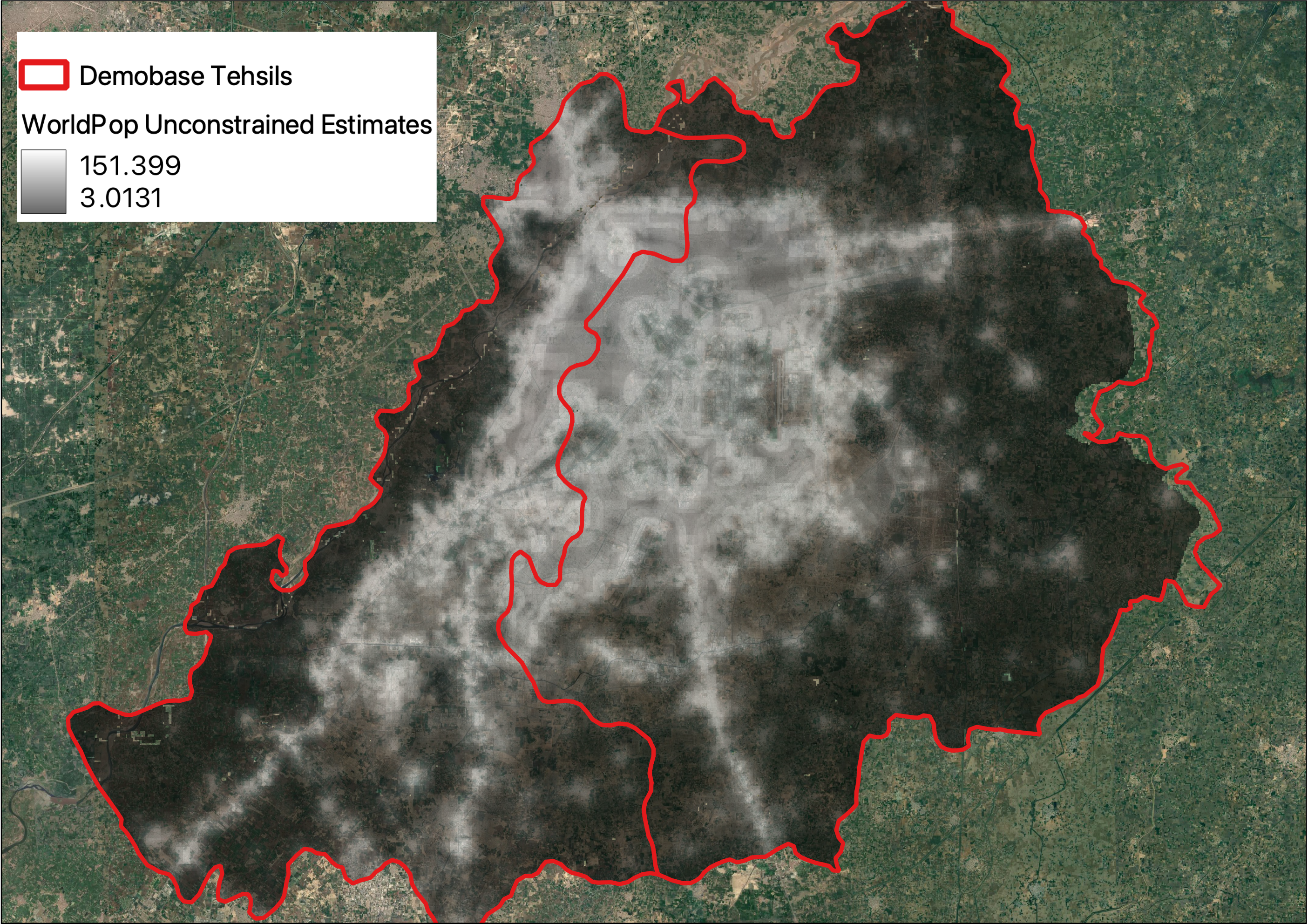}
  \label{Fig: 2(a)}
}
\subfloat[]{
  \includegraphics[width=\columnwidth, height = 60mm, keepaspectratio]{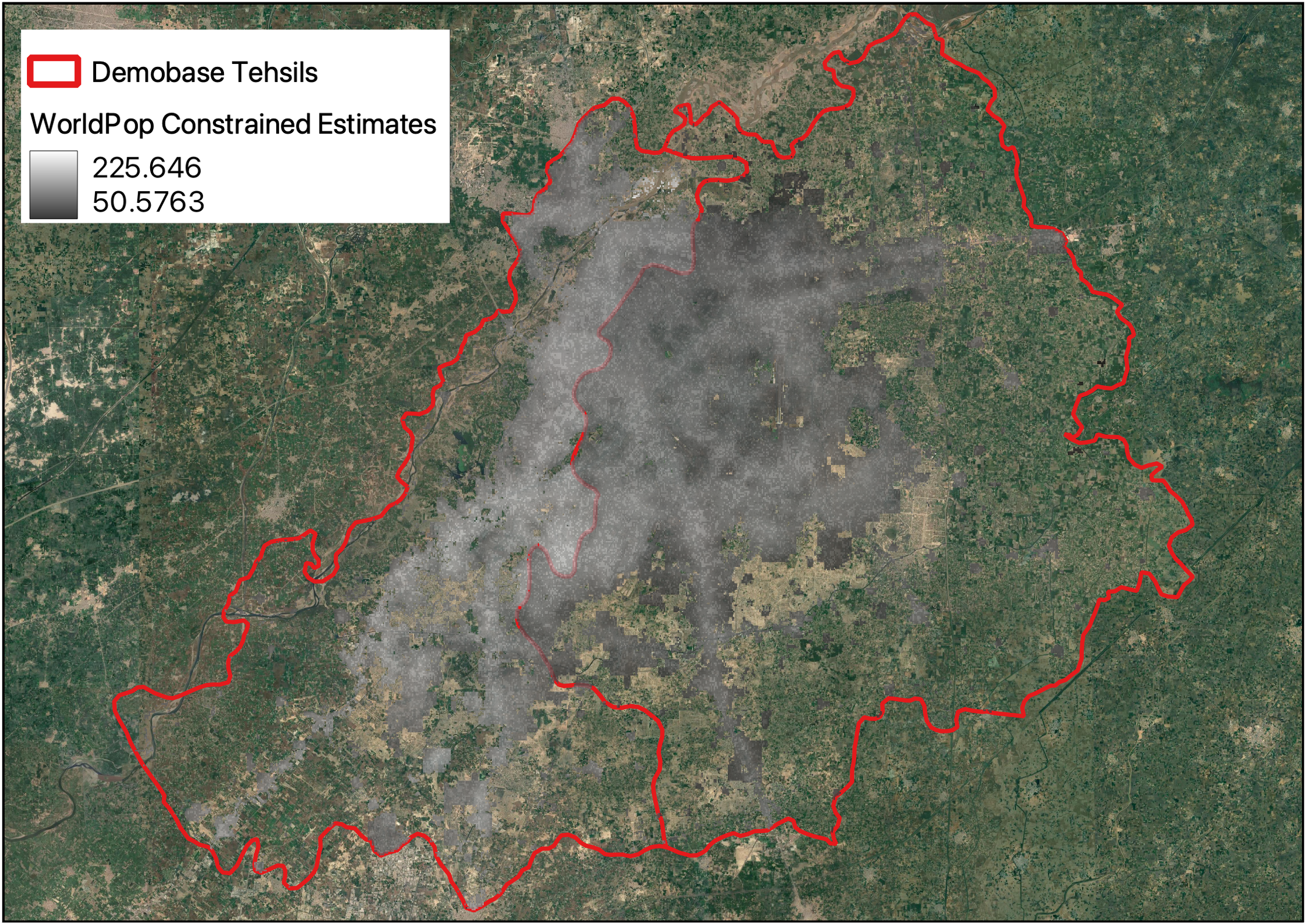}
  \label{Fig: 2(b)}
}
\vfill
\subfloat[]{
  \includegraphics[width=\columnwidth, height = 60mm, keepaspectratio]{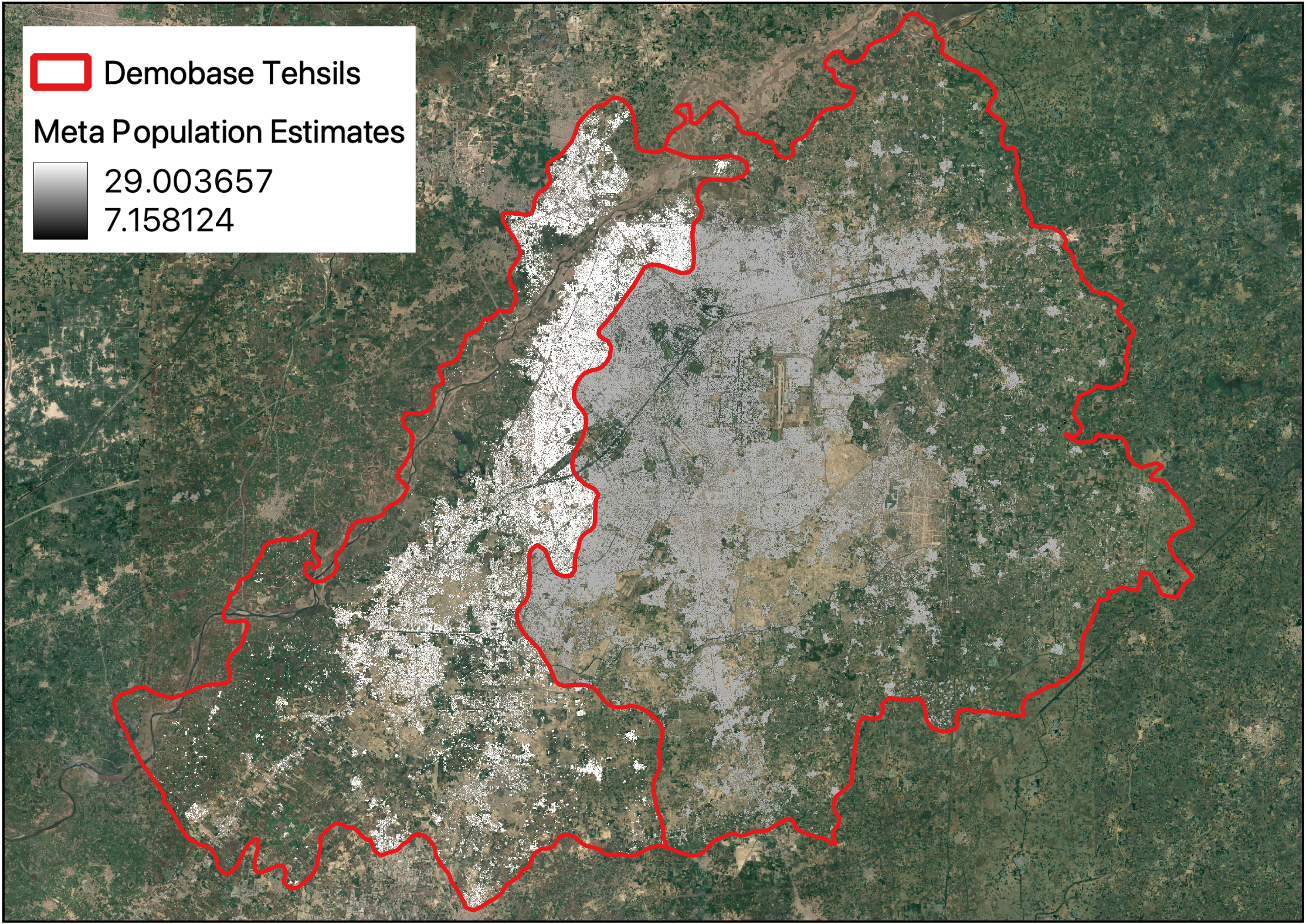}
  \label{Fig: 2(c)}
}
\subfloat[]{
  \includegraphics[width=\columnwidth, height = 60mm, keepaspectratio]{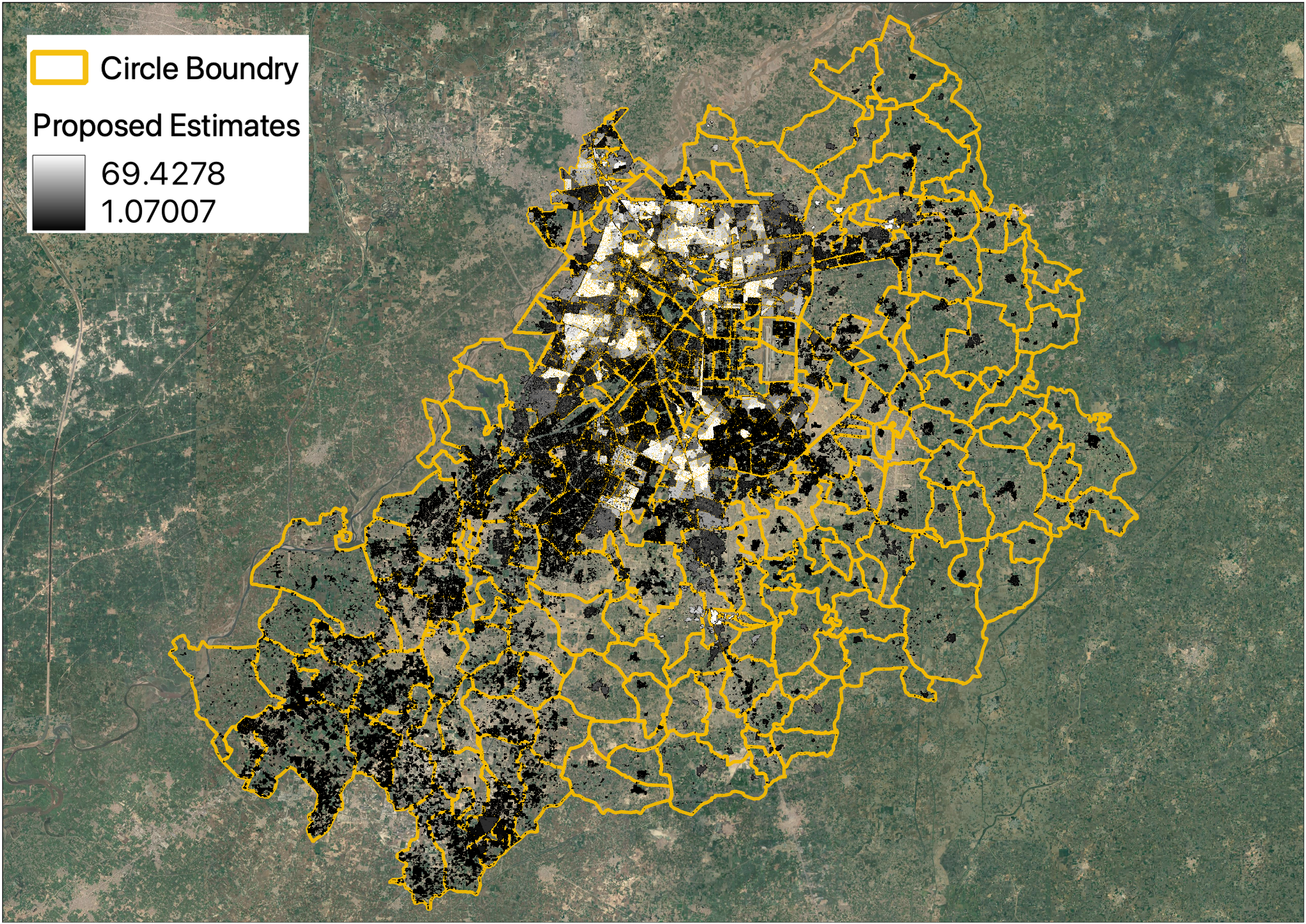}
  \label{Fig: 2(d}
}
\vfill
\caption{Population estimates of each dataset; (a) Worldpop Unconstrained, (b) Worldpop Constrained, (c) Meta and (d) Proposed}
\label{Fig: 2}
\end{figure*}
\section{Evaluation and Analysis}
\label{evaluation and analysis}
A comparison of our population density maps with those offered by Meta and WorldPop is shown in Fig. \ref{Fig: 2}. WorldPop unconstrained (a) tends to overestimate population counts by projecting population for agricultural and
barren land. WorldPop constrained (b) tries to overcome this shortcoming but also has severe limitations since it uses a built settlement growth model to determine the built-up area. The Meta (c) dataset effectiveness is limited since it uses the same population estimate for each tile in a tehsil. The proposed (d) method performs significantly better in estimating population counts than any other state-of-the-art methods since it uses up-to-date census data at the fourth administrative level along with precise constructed settlement masks. We are able to compute distinct population estimates for each tile using (2) and eliminate unbuilt regions by taking into consideration the proportion of built-up area in a tile to the total built-up area in a circle. Non-residential built-up tiles are removed using the POI data. The effectiveness of the created settlement mask is displayed in Table 1 for each of the aforementioned datasets.

\begin{table}[ht]
\centering
    \caption{Marked built-up areas are used to estimate accuracy and the F1-Score as percentages for the WorldPop constrained (WPC), Meta, and proposed datasets. The proposed beats cutting-edge datasets in every criterion, including the WPC dataset, which has a low resolution of 100 meters by 100 meters, demonstrating that it more consistently captures built-up areas.}
    \centering
    \label{tab:eval1}
    \begin{tabular}{|m{2.4cm}|m{2.4cm}|m{2.4cm}|}
    \hline
    \bf Dataset & \bf Accuracy & \bf F1-Score \\ \hline
    WPC  & 82.1 & 90.1\\ \hline 

    Meta  & 70.7 & 79.0 \\ \hline 
    
    Proposed  & \bf 86.7 & \bf 91.2  \\ \hline 

    \end{tabular}
\end{table}

\section{Conclusion}
\label{conclusion}
In order to more accurately estimate population counts at a 30-meter by 30-meter resolution, our method advances previous approaches by integrating more dependable built-settlement masks, latest census results and excluding non-residential areas. In addition, we plan to calculate the population of the entire country of Pakistan by comparing the population in each tile to the built-up area while accounting for the geographic diversity in Lahore's demography. We hope to help close the data gap in urban policy and research by making our high-resolution accurate population estimates available to the public. 
\bibliographystyle{IEEEbib}
\bibliography{strings,refs}

\end{document}